\documentclass{llncs}
\usepackage{booktabs} 
\usepackage{comment}
\usepackage{graphicx}
\usepackage[lofdepth,lotdepth]{subfig}
\usepackage{cellspace}
\usepackage{wrapfig,picins}
\usepackage[export]{adjustbox}

\usepackage[sorting=none, maxnames=50]{biblatex}
\usepackage{amsmath}
\usepackage{color}
\usepackage{url}
\usepackage{booktabs} 
\usepackage{amsfonts}
\DeclareGraphicsExtensions{.pdf,.png,.jpg}
\usepackage[english]{babel}
\usepackage[table,xcdraw]{xcolor}
\usepackage[section]{placeins}

\usepackage{listings}
\usepackage{xcolor}
\definecolor{codegreen}{rgb}{0,0.6,0}
\definecolor{codegray}{rgb}{0.5,0.5,0.5}
\definecolor{codepurple}{rgb}{0.58,0,0.82}
\definecolor{backcolour}{rgb}{0.95,0.95,0.92}
\lstdefinestyle{mystyle}{
    backgroundcolor=\color{backcolour},   
    commentstyle=\color{codegreen},
    keywordstyle=\color{magenta},
    numberstyle=\tiny\color{codegray},
    stringstyle=\color{codepurple},
    basicstyle=\ttfamily\footnotesize,
    breakatwhitespace=false,         
    breaklines=true,                 
    captionpos=b,                    
    keepspaces=true,                 
    numbers=left,                    
    numbersep=5pt,                  
    showspaces=false,                
    showstringspaces=false,
    showtabs=false,                  
    tabsize=2
}
\begin{document}

\frontmatter          
\pagestyle{empty}  

\title{A new method for binary classification of proteins with Machine Learning}
%
%
 \author{Damiano Perri\inst{1} $^{ORCID: 0000-0001-6815-6659}$\newline Marco Simonetti\inst{1} $^{ORCID: 0000-0003-2923-5519}$\newline Andrea Lombardi\inst{2} $^{ORCID:0000-0002-7875-2697}$\newline  Noelia Faginas-Lago\inst{2} $^{ORCID:0000-0002-4056-3364}$\newline Osvaldo Gervasi\inst{3} $^{ORCID: 0000-0003-4327-520X}$ 
 }
\institute{
University of Florence, Dept. of Mathematics and Computer Science, Florence, Italy \and University of Perugia, Dept. of Chemistry, Biology and Biotechnology, Perugia, Italy\and University of Perugia, Dept. of Mathematics and Computer Science, Perugia, Italy
}
\titlerunning{A new method for binary classification of proteins with Machine Learning} 
\authorrunning{D. Perri, M.  Simonetti, A. Lombardi, M. N. Faginas Lago and O. Gervasi} 

\maketitle

\begin{abstract}
In this work we set out to find a method to classify protein structures using a Deep Learning methodology. Our Artificial Intelligence has been trained to recognize complex biomolecule structures extrapolated from the Protein Data Bank (PDB) database and reprocessed as images; for this purpose various tests have been conducted with pre-trained Convolutional Neural Networks, such as InceptionResNetV2 or InceptionV3, in order to extract significant features from these images and correctly classify the molecule.
A comparative analysis of the performances of the various networks will therefore be produced.
\end{abstract}


\section{Introduction}

The classification of the geometric structures of proteins and the individuation of possible simple criteria to base their discrimination are complex tasks and a longstanding issue in chemical sciences. To investigate the relationships between structure and activity and for a satisfactory theoretical understanding of the protein folding process, the ability to assess the ``correctness'' and similarity of possible spatial arrangements of such macromolecules is a prerequisite.\\
The recent increased practicability of computational approaches based on Deep Learning and Neural Networks further motivates renewed efforts in such direction, since it permits one to resort to approaches based on the search for hidden patterns and regularities across large set of experimentally resolved protein structures.

In a recent paper \cite{perri2020binary}, we developed an approach to the basic problem of classifying as "real" a protein given its amino acid sequence, using a Deep Learning approach, based upon a Convolutional Neural Network (CNN) trained on a large set of data.\\
In the present paper, which is intended as a continuation of such previous work, a Convolutional Neural Network is again aimed at classifying as "true" or "false" a given structure, but the CNN has been developed after new significant improvements to the original approach for the recognition of the geometric structures. The idea was not to lose valuable spatial information regarding the shape of the protein to be examined, so we moved beyond the molecule model as a simple sequence of amino acids, to get to a more effective and realistic description preserving spatial information.
To this purpose we exploited the well known suitability of Convolutional Neural Networks for image analysis, where they are particularly appreciated in the recognition of images and their characteristics, both on two-dimensional or three-dimensional objects, through the extraction of particular features from images so that different kind of objects, like people or things, can be correctly classified. 

In this article we illustrate our approach to the problem using a two-dimensional representation based on 2D Convolutional Neural Networks, where any given protein is mapped into a two-dimensional grid of coloured pixels and then processed by the CNN in order to extract the relevant features and the characteristic properties to carry out the protein classification.
In order to train the neural network, similarly to the previous work, the set of protein structures was obtained from the Protein Data Bank (PDB)\cite{PDB} an open access repository containing data about proteins and nucleic acids' structures. 

The paper is organized as follows. In Sec.~\ref{classi} we briefly point out some key points about data extraction for protein classification. Sec.~\ref{method} illustrates the characteristics of our CNN, reports details about data extraction and processing. Preliminary applications, for training and validation, are also reported. Conclusions and perspectives are in Sec.~\ref{conclu}. 


\section{Strategies for classification of complex molecular structures}\label{classi}

For some years now, the use of Machine Learning techniques has rapidly become more and more pervasive in the world of Biology\cite{cartwright2008artificial, krishnan2003comparative} and Chemistry \cite{dral2020quantum, goldman2006machine, panteleev2018recent, mater2019deep}, especially in the field of classification of macro-molecules that are generally found in the modeling of protein and bio-molecular structures\cite{bakhtiarizadeh2014neural}.
The identification and correct assignment of the protein attribute to a generic bio-molecule is of considerable importance both for the purposes of genomic mapping and for the preparation of new and more specific groups of drugs\cite{ou2012computational, sliwoski2014computational, jamali2016drugminer}. \newline Various methodologies are continuously suggested that take into consideration different aspects\cite{roy2014selection}, such as chemical-physical properties \cite{taniguchi2020combination} or geometric structure \cite{brandt2018machine, cang2018representability}, to reach the goal.
In the first case, two different approaches are possible:
\begin{enumerate}
    \item attention is focused on a chemical-physical feature of interest and the bio-molecules are labelled, also using a deep learning or SVM (Support Vector Machine) technique to obtain an accurate and automated classifier;
    \item an n-dimensional vector with the descriptors of the chemical-physical properties to be examined is produced with an attached class label\cite{gromiha2008neural}.
\end{enumerate}
Several studies have confirmed that the combination of protein characteristics is preferable in order to obtain better predictive information than the use of single protein characteristics\cite{ong2007efficacy}.\newline
In the second case, the spatial arrangement of the various amino acid chains is evaluated for classification, both with SVM\cite{cai2003svm} and Machine Learning/Deep Learning\cite{cui2007advances, ding2001multi} techniques.
Today, convolutional neural networks are widely used in image analysis.
They allow the extraction of features thanks to which object recognition or image classification can be performed\cite{cnnDamianoDividiti2019,bocca2019,vella2021}.
Our research fits into this last channel with the aim of testing a way to correctly classify protein groups. Most of the work in this area has mainly focused on the study of the sequence and position of amino acids in the protein chain (primary or secondary sequence); our research has tried to maintain the information related to the sequences, simultaneously capturing all the geometric characteristics using 2D axonometric maps.

\section{Methodology: the architecture of the system}\label{method}
The system is a classic binary classifier whose task is to correctly subdivide the biomolecules given in input as "protein" or "non-protein". The data relating to the molecules to be examined are passed to the neural network as two-dimensional image processing, which faithfully reconstructs the geometric structure of the molecule itself.

\subsection{Data extraction and processing}

In order to validate our methodology, we have extracted from the PDB a sufficient number of records useful to effectively train the network; after several tests, it was found that good results were obtained with a number of samples around 3,000 units. This allowed us to select a group of proteins (equal to 2,911 molecules, with 16,924,350 amino acids) with the best images of their structure and focus our attention on the results.
The whole process, from data extraction to image evaluation by the neural network, was performed in Python3, with the help of well-known libraries useful for scientific computing and data processing, such as Numpy and Pandas.

\begin{figure}[ht]
    \centering
    \includegraphics[width=\linewidth]{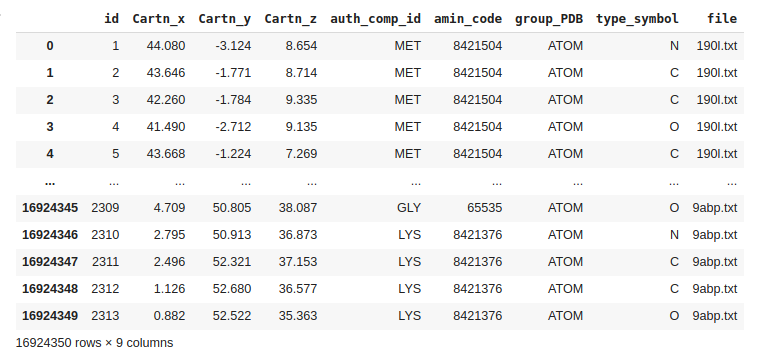}
    \caption{Information extracted: spatial position, name and colour code of amino acids}
    \label{fig01}
\end{figure}

 The various steps required for the generation and management of the dataset are listed below:
 \\ 1) Data extraction in XML format from the PDB database, identification of the necessary information we wish to keep and its subsequent transformation into a single object managed by the Pandas library (Fig.\ref{fig01})
\\ 2) Data cleaning, with the elimination of any duplicate records, possibly generated by the two different methods of measuring the crystal lattice structure for the molecule
\\ 3) Association of a unique RGB color code to each amino acid present in proteins (e.g. Alanine 128, Glycine 65280, Lysine 8421376), in order to visualize the structure of the molecule as an image, on which every single amino acid is coloured in a different way
\begin{figure}[ht]
    \centering
    \includegraphics[width=\linewidth]{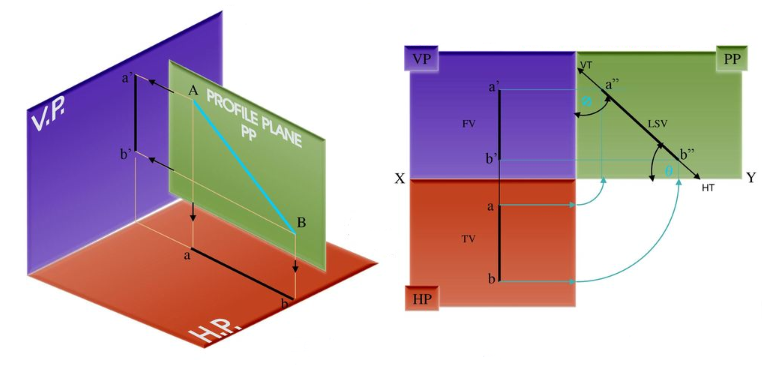}
    \caption{Orthogonal axonometry}
    \label{fig02}
\end{figure}
\begin{itemize}
    \item Visualization of the molecule according to a multiview orthographic axonometric map with orthogonal projections on three planes (horizontal plane, vertical plane and lateral/profile plane - an example of this type of axonometry is shown in Figure \ref{fig02}), in order to split and project the whole 3D image on a flat surface, without losing the isometricity and symmetries on the x, y and z components for the individual amino acids. From the analysis of the coordinates of all the amino acids present in the dataset, it was possible through an appropriate translation for the origin of the Cartesian reference system and an integer mapping for the numerical values of the coordinates themselves to represent each single protein in the domain $D=[0,3200]^3 \subset \mathbb{N}^3$; this allowed us to refer to each point (x, y, z) belonging to the cube D as a 3-indices tuple for the tensor with dimensions 3200x3200x3200, capable of containing the entire biomolecule
    \item Each image has been processed to fit within the 299x299 pixel dimensions, necessary as input dimensions for a 2D convolutional neural network; in order to avoid distorting the original axonometric proportions, the figures have been carefully cut out at the edges and in the central areas, to reduce unnecessary black padding.
\end{itemize}
\begin{figure}[ht]
    \subfloat[Subfigure 1 list of figures text][5AFR, N-terminal fragment of dynein heavy chain]{
    \includegraphics[width=0.5\linewidth]{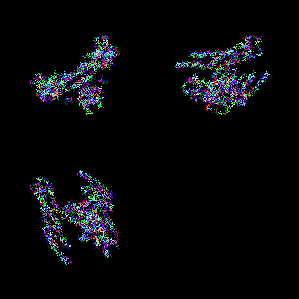}
    \label{fig:subfig1}}
    \qquad
    \subfloat[Subfigure 2 list of figures text][5AGU, sliding clamp of Mycobacterium tuberculosis]{
   \includegraphics[width=0.5\linewidth]{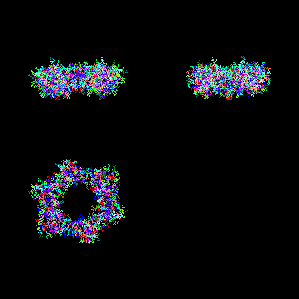}
   \label{fig:subfig2}}
    \qquad
    \subfloat[Subfigure 3 list of figures text][6ABO, human XRCC4 and IFFO1 complex]{
    \includegraphics[width=0.5\linewidth]{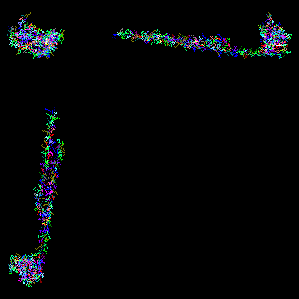}
    \label{fig:subfig3}}
    \qquad
    \subfloat[Subfigure 4 list of figures text][6AGX, cocrystal structure of FGFR2]{
    \includegraphics[width=0.5\linewidth]{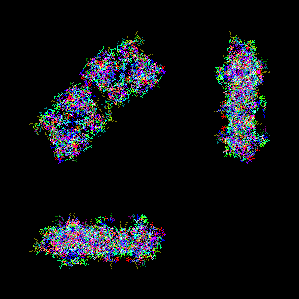}
    \label{fig:subfig4}}
    \qquad
 \caption{Images of 4 proteins obtained with our representation method}
 \label{figure:proteins}
\end{figure}
Therefore, for each single protein, the various amino acids were projected, as colored dots, on the three main planes: in Figure \ref{figure:proteins} four images obtained with our method for four different proteins are shown.\\
4) Generation of false samples, necessary for the learning of the neural network. It was decided to proceed starting from the original images; for the single amino acids belonging to each protein we applied a mutation probability, established at the beginning of the process (in our case experimental tests made us lean towards a fixed value of 5\%), which induced a colour change on the coloured points representing the amino acid: 2911 images of false proteins were thus produced. In Fig. \ref{fig04} a portion of a true protein is reproduced (above) with its false analogue (below): it is possible to notice the differences due to the mutation process (pixel that occupies the same position in the two figures, but has different colors).

\begin{figure}[ht]
    \centering
    \includegraphics[width=\linewidth]{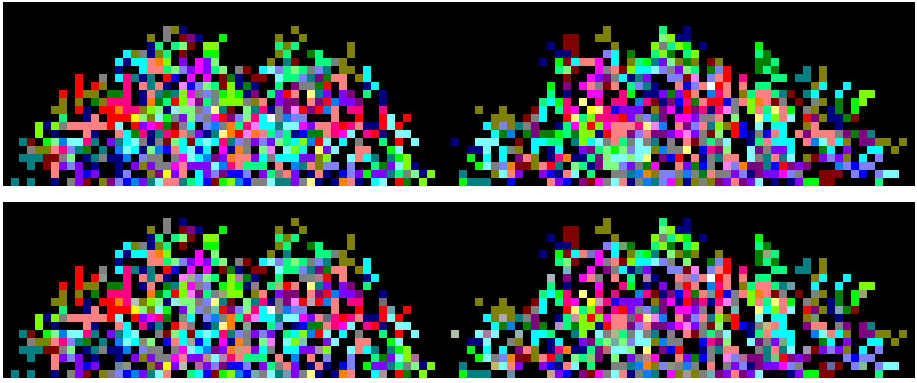}
    \caption{Differences between the same portion of a real molecule (above) and its false analogue (below)}
    \label{fig04}
\end{figure}

\subsection{Training and validation}\label{result}
The performances of two neural networks were analyzed: InceptionV3 and InceptionResNetV2.
These networks were trained through the \textbf{transfer learning} technique which allowed us to obtain high accuracy values with a reduced number of learning epochs.\newline
Transfer learning is the process by which a neural network learn a new task through knowledge's transfer a related task previously learned. Using transfer learning, the weight values of the convolutional layers are imported (they are generally used for feature extraction and already present as initial parameters of the network itself), and only the final layers of the neural network that are generally used for classification are trained\cite{perriFranzoni,cancroPelle}.
Both networks are pre-trained on ImageNet which is an image dataset used for object recognition consisting of 14 million photographs.\newline
The model used for the InceptionV3 neural network is shown in the table \ref{table:inceptionv3}, while the model used for the InceptionResNetV2 neural network is shown in the table \ref{table:inceptionresnetv2}.

\begin{table}[ht]
\centering
\begin{tabular}{ccc}
\hline
\multicolumn{1}{|c|}{Layer (type)} & \multicolumn{1}{c|}{\cellcolor[HTML]{FFFFFF}Output Shape} & \multicolumn{1}{c|}{\cellcolor[HTML]{FFFFFF}Param \#} \\ \hline
\rowcolor[HTML]{FFFFFF} 
\multicolumn{1}{|c|}{\cellcolor[HTML]{FFFFFF}inception\_v3} & \multicolumn{1}{c|}{\cellcolor[HTML]{FFFFFF}(None, 8, 8, 2048)} & \multicolumn{1}{c|}{\cellcolor[HTML]{FFFFFF}21802784} \\ \hline
\rowcolor[HTML]{FFFFFF} 
\multicolumn{1}{|c|}{\cellcolor[HTML]{FFFFFF}flatten (Flatten)} & \multicolumn{1}{c|}{\cellcolor[HTML]{FFFFFF}(None, 131072)} & \multicolumn{1}{c|}{\cellcolor[HTML]{FFFFFF}0} \\ \hline
\rowcolor[HTML]{FFFFFF} 
\multicolumn{1}{|c|}{\cellcolor[HTML]{FFFFFF}dense (Dense)} & \multicolumn{1}{c|}{\cellcolor[HTML]{FFFFFF}(None, 64)} & \multicolumn{1}{c|}{\cellcolor[HTML]{FFFFFF}8388672} \\ \hline
\multicolumn{1}{|c|}{dense\_1 (Dense)} & \multicolumn{1}{c|}{(None, 64)} & \multicolumn{1}{c|}{4160} \\ \hline
\multicolumn{1}{|c|}{dense\_2 (Dense)} & \multicolumn{1}{c|}{(None, 1)} & \multicolumn{1}{c|}{65} \\ \hline
\multicolumn{3}{c}{=====================================} \\ \hline
\multicolumn{2}{|c|}{Total Params:} & \multicolumn{1}{c|}{30,195,681} \\ \hline
\multicolumn{2}{|c|}{Trainable Params:} & \multicolumn{1}{c|}{30,161,249} \\ \hline
\end{tabular}\caption{InceptionV3 model}\label{table:inceptionv3}
\end{table}

\begin{table}[]
\centering
\begin{tabular}{
>{\columncolor[HTML]{FFFFFF}}c 
>{\columncolor[HTML]{FFFFFF}}c 
>{\columncolor[HTML]{FFFFFF}}c }
\hline
\multicolumn{1}{|c|}{\cellcolor[HTML]{FFFFFF}Layer (type)} & \multicolumn{1}{c|}{\cellcolor[HTML]{FFFFFF}Output Shape} & \multicolumn{1}{c|}{\cellcolor[HTML]{FFFFFF}Param \#} \\ \hline
\multicolumn{1}{|c|}{\cellcolor[HTML]{FFFFFF}inception\_resnet\_v2} & \multicolumn{1}{c|}{\cellcolor[HTML]{FFFFFF}(None, 8, 8, 1536)} & \multicolumn{1}{c|}{\cellcolor[HTML]{FFFFFF}54336736} \\ \hline
\multicolumn{1}{|c|}{\cellcolor[HTML]{FFFFFF}flatten (Flatten)} & \multicolumn{1}{c|}{\cellcolor[HTML]{FFFFFF}(None, 98304)} & \multicolumn{1}{c|}{\cellcolor[HTML]{FFFFFF}0} \\ \hline
\multicolumn{1}{|c|}{\cellcolor[HTML]{FFFFFF}dense (Dense)} & \multicolumn{1}{c|}{\cellcolor[HTML]{FFFFFF}(None, 64)} & \multicolumn{1}{c|}{\cellcolor[HTML]{FFFFFF}6291520} \\ \hline
\multicolumn{1}{|c|}{\cellcolor[HTML]{FFFFFF}dense\_1 (Dense)} & \multicolumn{1}{c|}{\cellcolor[HTML]{FFFFFF}(None, 64)} & \multicolumn{1}{c|}{\cellcolor[HTML]{FFFFFF}4160} \\ \hline
\multicolumn{1}{|c|}{\cellcolor[HTML]{FFFFFF}dense\_2 (Dense)} & \multicolumn{1}{c|}{\cellcolor[HTML]{FFFFFF}(None, 1)} & \multicolumn{1}{c|}{\cellcolor[HTML]{FFFFFF}65} \\ \hline
\multicolumn{3}{c}{\cellcolor[HTML]{FFFFFF}=====================================} \\ \hline
\multicolumn{2}{|c|}{\cellcolor[HTML]{FFFFFF}Total Params:} & \multicolumn{1}{c|}{\cellcolor[HTML]{FFFFFF}60,632,481} \\ \hline
\multicolumn{2}{|c|}{\cellcolor[HTML]{FFFFFF}Trainable Params:} & \multicolumn{1}{c|}{\cellcolor[HTML]{FFFFFF}60,571,937} \\ \hline
\end{tabular}\caption{InceptionResNetV2 model}\label{table:inceptionresnetv2}
\end{table}

\begin{figure}[ht]
    \subfloat[][InceptionV3: Training set]{
    \includegraphics[trim=100 13 40 24 ,clip,width=0.5\linewidth]{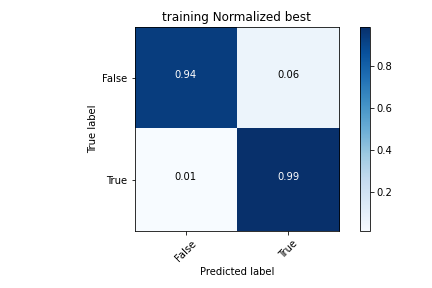}
    \label{fig:subfig2.1}}
    \qquad
    \subfloat[][InceptionV3: Validation set]{
   \includegraphics[trim=100 13 40 24 ,clip,width=0.5\linewidth]{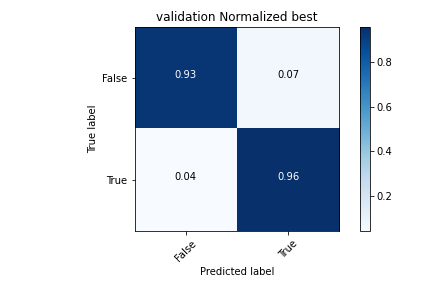}
   \label{fig:subfig2.2}}
    \qquad
    \subfloat[][InceptionResNetV2: Training set]{
    \includegraphics[trim=100 13 40 24 ,clip,width=0.5\linewidth]{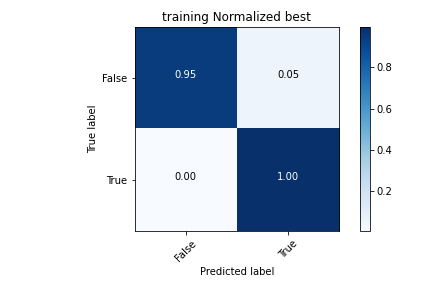}
    \label{fig:subfig2.3}}
    \qquad
    \subfloat[][InceptionResNetV2: Validation set]{
    \includegraphics[trim=100 13 40 24 ,clip,width=0.5\linewidth]{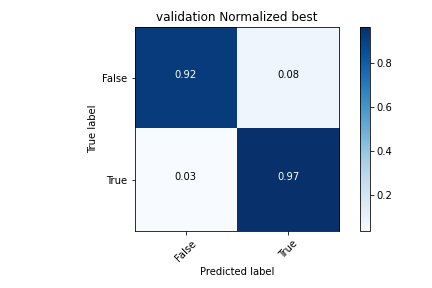}
    \label{fig:subfig2.4}}
    \qquad
 \caption{Confusion matrices}
 \label{figure:confusionMatrices}
\end{figure}

The results of the training are summarized in the table \ref{table1:results}, where it can be seen that the InceptionV3 network has a higher accuracy's percentage in recognizing false proteins from true ones compared to the results obtained by InceptionResNetV2.
\begin{table}[ht]
\centering
\begin{tabular}{
>{\columncolor[HTML]{FFFFFF}}c |
>{\columncolor[HTML]{FFFFFF}}c |
>{\columncolor[HTML]{FFFFFF}}c |}
\cline{2-3}
 & InceptionV3 & InceptionResNetV2 \\ \hline
\multicolumn{1}{|c|}{\cellcolor[HTML]{FFFFFF}Accuracy Train Set} & 99.06\% & 98.36\% \\ \hline
\multicolumn{1}{|c|}{\cellcolor[HTML]{FFFFFF}Accuracy Test Set} & 94.57\% & 92.14\% \\ \hline
\multicolumn{1}{|c|}{\cellcolor[HTML]{FFFFFF}Model Size} & 461MB & 465MB \\ \hline
\end{tabular}\caption{Final Results}\label{table1:results}
\end{table}
The Confusion Matrices\cite{visa2011confusion} of the two models indicate that they are both able to classify proteins with good results, also we believe that a correct parameters' modulation, through the realization of ad-hoc models, can further increase the degree of accuracy.
The matrices are shown in figure \ref{figure:confusionMatrices}.

\section{Conclusions and future works}\label{conclu}
In this work we presented an improved Convolutional Neural Network for the classification of protein structure. Preliminary results appear encouraging, showing that the methodology used for the representation of the protein geometry, based on a 2D projected image associated to each molecule, retains most of the spatial information and is suitable for recognition. Also, the computational performances seems quite good,
being calculations extremely fast as we provide the neural network with an image for each molecule.

However, considerations on the general nature of the CNNs used lead us to think that specifically designed neural networks could significantly improve the results, or even outperform them.
A further research path worth being followed is to train the neural network using a greater number of samples, to better analyze the link between the samples' structural complexity and the classification capacity of the neural network itself.
We believe that even better performance might be achieved if we developed a neural network customised for the graphical representation we proposed.
Our representation is in fact very particular and within our images the areas with a high information content are located in very specific sectors of the images.
Furthermore, a personalised neural network could also reduce the size in MegaBytes of the model obtained in output.

\section{Acknowledgments}
AL and NFL thank the Dipartimento di Chimica, Biologia e Biotecnologie dell'Universit\`{a} di Perugia
(FRB, Fondo per la Ricerca di Base 2019 and 2020) and the
Italian MIUR and the University of Perugia for the financial support of the AMIS project through the program ``Dipartimenti di
Eccellenza''.
AL thanks the OU Supercomputing Center for Education \& Research
(OSCER) at the University of Oklahoma, for allocated
computing time.

%
%
\printbibliography
\newpage
\end{document}